%% file: arxiv_main.tex
\documentclass{article} 
\usepackage{iclr2026_conference,times}

\input{math_commands.tex}

\usepackage{hyperref}
\usepackage{url}

\usepackage{booktabs}
\usepackage{graphicx}
\usepackage{wrapfig}
\usepackage{multicol}
\usepackage{multirow}
\usepackage{subcaption}
\usepackage[table]{xcolor}

\colorlet{highlight}{cyan!10}

\title{Do LLMs Signal When They’re Right?\\Evidence from Neuron Agreement}

\author{Kang Chen$^{1\dagger}$,
    Yaoning Wang$^{1\dagger}$,
    Kai Xiong$^{2}$,
    Zhuoka Feng$^{1}$,
    Wenhe Sun$^{1}$,\\
    \textbf{Haotian Chen$^{1}$, Yixin Cao$^{1}$}\thanks{Corresponding Author\\ \hspace*{1.35em}$^\dagger$Equal Contribution}\\
    \textsuperscript{1}Institute of Trustworthy Embodied AI, Fudan University\\
    \textsuperscript{2}Harbin Institute of Technology\\
    \texttt{Kchen24@m.fudan.edu.cn, yxcao@fudan.edu.cn}
}

\iclrfinalcopy
\begin{document}

\maketitle

\begin{abstract}
Large language models (LLMs) commonly boost reasoning via sample-evaluate-ensemble decoders, achieving label free gains without ground truth. However, prevailing strategies score candidates using only external outputs such as token probabilities, entropies, or self evaluations, and these signals can be poorly calibrated after post training. We instead analyze internal behavior based on neuron activations and uncover three findings: (1) external signals are low dimensional projections of richer internal dynamics; (2) correct responses activate substantially fewer unique neurons than incorrect ones throughout generation; and (3) activations from correct responses exhibit stronger cross sample agreement, whereas incorrect ones diverge. Motivated by these observations, we propose Neuron Agreement Decoding (NAD), an unsupervised best-of-N method that selects candidates using activation sparsity and cross sample neuron agreement, operating solely on internal signals and without requiring comparable textual outputs. NAD enables early correctness prediction within the first 32 generated tokens and supports aggressive early stopping. Across math and science benchmarks with verifiable answers, NAD matches majority voting; on open ended coding benchmarks where majority voting is inapplicable, NAD consistently outperforms Avg@64. By pruning unpromising trajectories early, NAD reduces token usage by 99\% with minimal loss in generation quality, showing that internal signals provide reliable, scalable, and efficient guidance for label free ensemble decoding.

\end{abstract}

\input{chapters_arxiv/intro}
\input{chapters_arxiv/related}
\input{chapters_arxiv/preliminaries}
\input{chapters_arxiv/method}
\input{chapters_arxiv/experiments}
\input{chapters_arxiv/conclusion}

\bibliography{iclr2026_conference}
\bibliographystyle{iclr2026_conference}

\input{chapters_arxiv/appendix}

\end{document}

%% file: math_commands.tex
\usepackage{amsmath,amsfonts,bm}

\def\eqref#1{equation~\ref{#1}}

\def\1{\bm{1}}

\DeclareMathAlphabet{\mathsfit}{\encodingdefault}{\sfdefault}{m}{sl}
\SetMathAlphabet{\mathsfit}{bold}{\encodingdefault}{\sfdefault}{bx}{n}



%% file: chapters_arxiv/intro.tex
\section{Introduction}
\label{sec:intro}
Large language models (LLMs) have demonstrated remarkable reasoning capabilities~\citep{wei2022chain,tang2024mathscale}. To further boost their performance, sample-evaluate-ensemble methods have been widely adopted. These methods leverage answer consistency by selecting the best response through majority voting~\citep{Wang2022SelfCons}, often yielding higher quality than single responses. Notably, this approach requires no ground truth labels, essentially providing a ``free lunch'' improvement. Beyond inference-time applications, this paradigm has also been integrated into unsupervised reinforcement learning, enabling models to train at larger scales without ground truth, thereby raising the performance ceiling.

Beyond majority voting, recent sample-evaluate-ensemble methods have developed more sophisticated ensemble strategies. Instead of treating all responses as equally important like majority voting,~\cite{Chen2023USC} prompts the model to self-evaluate before ensemble and select the most consistent answer among candidates. Another line of work explores the model's response confidence, leveraging output states from the forward pass (e.g., token probabilities) to evaluate and ensemble responses. Typically, high-confidence responses exhibit better quality and low-confidence ones shall be pruned~\citep{Fu2025DeepConf}. However, these methods rely solely on model outputs, with confidence or entropy metrics based on token probabilities --- what we define as the external behaviors of LLMs. This raises critical questions: Do models inherently encode response quality signals in their outputs? How reliable are such assessments? According to GPT-4 reports, LLMs lose calibration capabilities after post-training, showing no clear linear relationship between token probability and response correctness. This appears to contradict the effectiveness demonstrated in existing work.

\begin{wrapfigure}{r}{0.5\textwidth}
  \centering
  \includegraphics[width=0.48\textwidth]{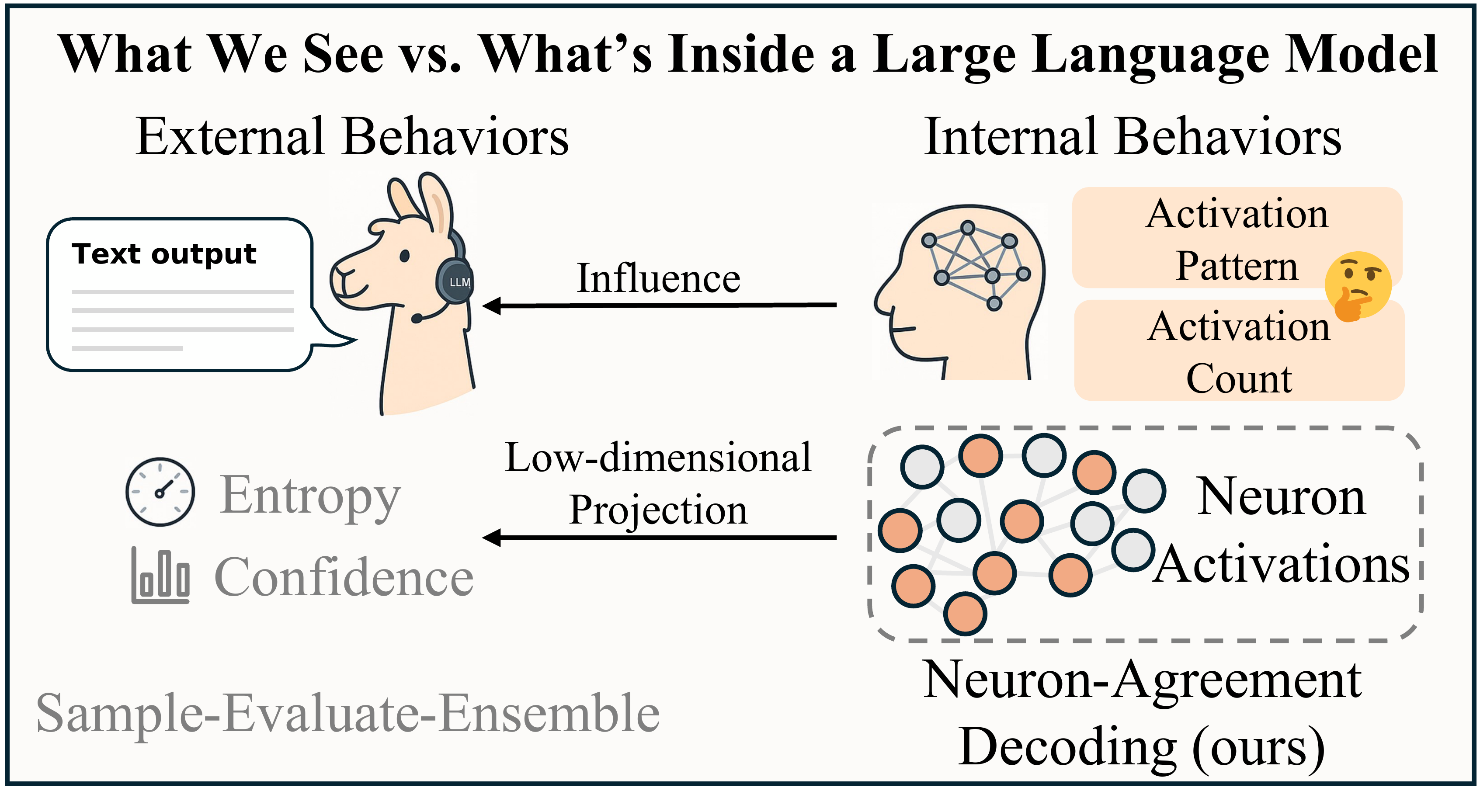} 
  \caption{Comparison between NAD and other ensemble methods. Our approach relies solely on internal signals during the sampling process, without requiring comparable textual outputs.}
  \label{fig:difference}
\end{wrapfigure}

In this paper, we further investigate the model's internal behaviors, neuron activation vectors, to explore their relationships with external behaviors and response correctness. Through extensive experiments, we reveal that: 1) External behaviors (e.g., entropy) represent low-dimensional projections of internal behaviors. This is intuitive, as token probabilities determine output tokens and are themselves determined by neuron activations across layers. 2) Consequently, internal behaviors contain richer signals, leading to our discovery of their relationship with response quality --- correct responses activate significantly fewer neurons compared to incorrect ones. 3) Finally, through visualization, we observe patterns among correct responses. They tend to activate similar unique neurons. These novel patterns in LLMs' internal behaviors inspire better evaluation methods and more efficient assessment of sampled response correctness, ultimately yielding superior ensemble results.

Building on these insights, we propose \textbf{Neuron-Agreement Decoding~(NAD)}, an unsupervised method that selects high-quality reasoning trajectories solely based on internal neuron activations. Specifically, NAD favors either responses with minimal neuron activations or those that exhibit the greatest agreement with other sampled responses. Leveraging such signals also enables us to predict response correctness at a very early stage of generation~(e.g., within the first 32 tokens), rather than requiring full sequences as in external ensemble methods, thereby substantially improving the efficiency and effectiveness of sample-then-ensemble decoding. The distinction between our approach and existing ensemble methods is illustrated in Figure~\ref{fig:difference}.

For evaluation, we consider tasks with easily verifiable correctness~(with ground truth, suitable for majority voting) and more challenging scenarios~(without ground truth or with multiple valid solutions, e.g., code generation), validating our method's effectiveness. Moreover, when combined with an early-stopping strategy, NAD reduces token consumption in parallel sampling by up to two orders of magnitude while delivering superior performance. Our contributions are summarized as follows:

\begin{itemize}
    \item We conduct a deep investigation into the relationships among internal neuron activation, external behaviors, and response correctness in LLMs.
    
    \item We design a Neuron-Agreement Decoding~(NAD) method for scalable best-of-N sampling.
    
    \item Experimental results validate the effectiveness of our findings and method, while the efficiency gains are also achieved through the early-stopping strategy.
\end{itemize}

%% file: chapters_arxiv/related.tex
\section{Related Work}
\paragraph{Outputs-based Voting.} Self-consistency~\citep{Wang2022SelfCons} is a test-time ensemble technique that samples multiple chain-of-thought (CoT) solutions from an LLM and selects the final answer by majority vote. This method significantly improved performance on arithmetic and commonsense QA benchmarks, revealing that while any single chain might be incorrect, aggregating multiple solutions can correct errors. Variants of this idea include Soft Self-Consistency~\citep{Wang2024SoftSC}, which gives partial credit to similar answers rather than exact match voting. Self-consistency works well when outputs are relatively constrained (e.g., numerical or factual answers), but is less applicable to open-ended generation, where answers cannot be easily compared for voting.

\paragraph{Confidence-Based Selection.} Another line of work exploits internal confidence or uncertainty metrics. \cite{Kang2025SelfCert} introduced self-certainty, which uses the model's token-level probabilities to estimate the confidence of each reasoning chain. In their best-of-$N$ selection framework, self-certainty scores guided the choice of the final answer, achieving better scaling with $N$. \cite{Fu2025DeepConf} proposed DeepConf, which monitors token prediction entropy during generation to prune low-confidence reasoning paths on the fly, thereby saving computation while maintaining accuracy. These approaches require access to the model's probability distribution at each step. However, these approaches still rely on the availability of comparable answers, which limits their applicability.

%% file: chapters_arxiv/preliminaries.tex
\section{Pilot Study of Internal Behaviors}
Existing ensemble methods still rely heavily on external signals, overlooking internal dynamics in LLMs. Inspired by the model utility law~\citep{cao2025model}, in this section, we investigate how internal signals in the model~(i.e., neuron activations) correlate with these external signals~(e.g. certainty and entropy), and whether we can leverage them to guide the selection of the highest-quality trajectory in the early stage of generation. In Section~\ref{sec:neuron}, we first introduce the definition of activated neurons; in Section~\ref{sec:correlation}, we demonstrate the correlation between neuron activations and external signals; in Section~\ref{sec:pre_experiment}, we present preliminary experiments about the correlation between neuron activation patterns and model performance, which highlight two key insights to serve as the main basis for our proposed method.

\subsection{Neuron Activation Set}
\label{sec:neuron}
Neuron activation patterns can reveal the internal dynamics of LLMs, which are closely associated with specific types of abilities in LLMs~\citep{pan-etal-2024-finding, templeton2024scaling}. Here, we adopt the definition of activated neurons of~\cite{cao2025model}: Given an LLM with an input $x$, it can generate an output token sequence $\bm{y}=(y_1,y_2,\dots,y_t)$ from a single sampling, the SwiGLU-based~\citep{chowdhery2023palm} contribution of neuron $i$ in layer $l$ to output token $y_j$ as:
\begin{equation}
\begin{aligned}
f_{\text{neuron}}(i,l, y_{j} \mid \bm{y}_{<j}) 
= \big(\mathbf{W}_u \mathbf{W}_{\text{out}}^l \circ
       \operatorname{SiLU}\!\big(\mathbf{x}_t^l \mathbf{W}_g^l\big)\big)_{y_{j},i},
\end{aligned}
\label{eq:neuron_contribution}
\end{equation}

where $\bm{y}_{<j} = (y_1,\,y_2,\,\ldots,\,y_{j-1})$ denotes the response sequence before the $j$-th token $y_j$, $\operatorname{SiLU}$ is the Swish activation~\citep{shazeer2020glu}, $\mathbf{W}_{\text{out}}^{l}, \mathbf{W}_{g}^{l}$ are the output/gate projections in FFN, $\mathbf{W}_u$ is the unembedding matrix transforming the hidden states into distributions over the vocabulary, $\circ$ is an element-wise product with broadcasting, and $\mathbf{x}^l_{j-1}$ denotes the hidden input of token $y_{j-1}$ to the FFN at $l$-th layer.
For a given threshold $\eta$, the activated neuron set for a sample $(x,\bm{y})$ is defined as: 

\begin{equation}
N_{\text{activated}}(x,\bm{y})
= \Bigl\{
(i, l)
\;\Big|\;
\exists y_j \in \bm{y}, f_{\text{neuron}}\left(i, l,\,y_j \mid x \oplus y_{<j}\right) > \eta 
\Bigr\},
\label{eq:neuron_contribution_case}
\end{equation}
where $l=1, 2, ...,L$ represents the layer index, and $i=1,2,...,N$ indicates the neuron index in each layer. The implementation of the threshold function $\eta$ can be found in Appendix~\ref{sec:threshold}.

In this work, we refine the definition into a more fine-grained form to better capture activations throughout the reasoning process. Specifically, we divide the entire reasoning process into $B$-sized chunks $\bm{y}=(\bm{y}_1,\bm{y}_2, ...,\bm{y}_{\lceil t/B \rceil})$, compute the activation set for each chunk using the definition in Eq.(\ref{eq:neuron_contribution_case}), and then take their union:
\begin{equation}
\label{eq:neuron_chunked}
    N_\text{activated}(x,\bm{y})=\bigcup_{i=1}^{\lceil t/B \rceil}N_\text{activated}(x,\bm{y}_i)
\end{equation}

This modification allows us to better capture localized information within the reasoning process.

\subsection{Neuron Activations: Approximate preimages of confidence metrics}
\label{sec:correlation}
We hypothesize (operationally) that a model’s internal dynamics during inference—captured as the set of activated neurons—can be viewed as giving rise to \textbf{approximate} low-dimensional summaries such as self-certainty; reducing behavior to such scalars discards much of the underlying high-dimensional information. We test this via two complementary findings: (1) \textbf{activated neurons can, to some extent, aggregate into those scalar metrics}, i.e., the set of activated neurons contains the information needed to compute those metrics; and (2) \textbf{activated neurons exhibit patterns that scalar metrics cannot capture}.

\begin{figure}[tb]
    \centering
    \includegraphics[width=0.65\textwidth]{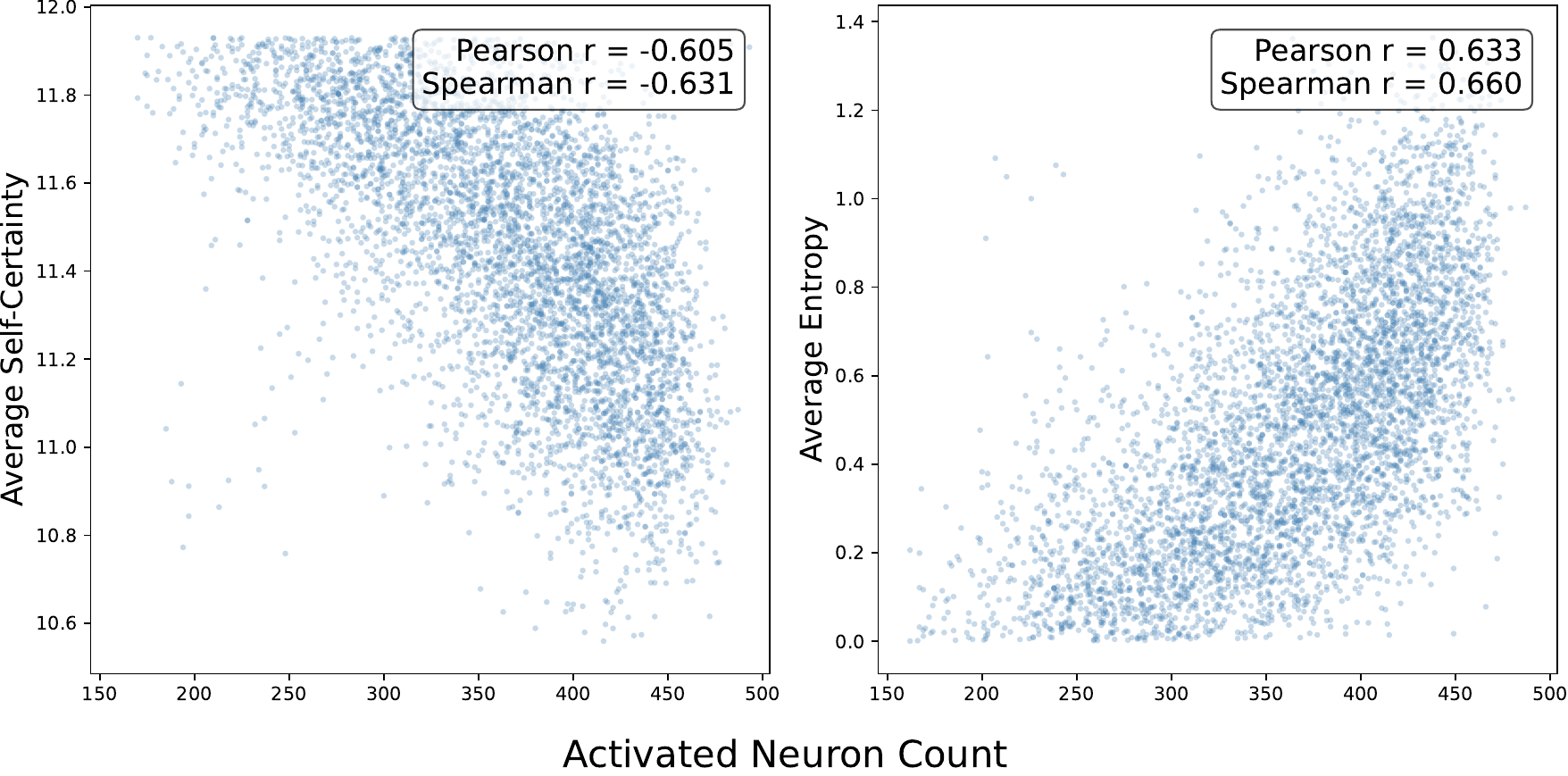}
    \caption{Scatter plots of the number of activated neurons versus confidence-based metrics (Self-Certainty and Entropy). The neuron counts show significant correlations with both metrics, indicating that the activated neuron states provide a high-dimensional representation of traditional confidence measures.}
    \label{fig:neuron_corr}
\end{figure}

\begin{wrapfigure}{R}{0.45\textwidth}
  \centering
  \includegraphics[width=0.4\textwidth]{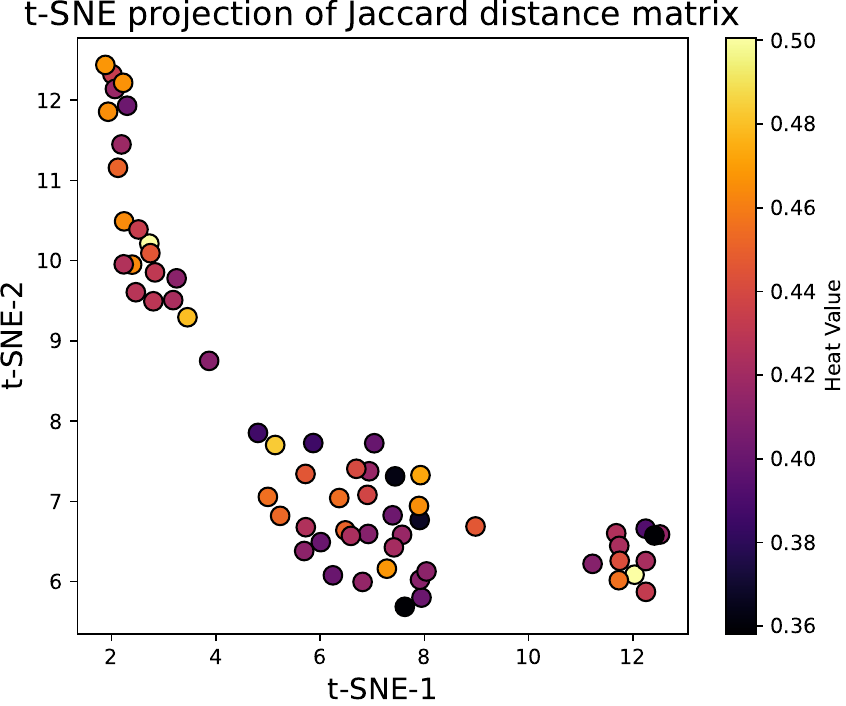} 
  \caption{t-SNE representation of activated neurons, with point colors indicating the average entropy of the corresponding samples. No clear consistency is observed between entropy and the resulting clusters, suggesting that the activated neurons contain high-dimensional structural information not captured by entropy.}
  \label{fig:neuron_entropy}
\end{wrapfigure}

Concretely, we use \texttt{Qwen3-4B-Think} to generate 64 responses per instance on \texttt{AIME24}, recording for each response both several scalar metrics and its corresponding set of activated neurons. For (1), we count the number of activated neurons per chunk and compute each chunk’s mean self-certainty and entropy, then measure the correlation between neuron counts and the conventional scalar metrics. The results, shown in Fig.~\ref{fig:neuron_corr}, demonstrate significant correlations (with p-value less than 0.05): neuron count correlates positively with entropy and negatively with self-certainty, thereby supporting (1). For (2), to uncover structure in how different samples activate neurons, we embed samples with t-SNE using the Jaccard index between activated-neuron sets as the similarity measure, i.e.
\begin{equation}
    S_{ij}=\frac{|N_\text{activated}(x,\pmb{y}_i)\bigcap N_\text{activated}(x,\pmb{y}_j)|}{|N_\text{activated}(x,\pmb{y}_i)\bigcup N_\text{activated}(x,\pmb{y}_j)|}.
\end{equation}
Where $\pmb{y}_i, \pmb{y}_j$ are different responses. The t-SNE visualization in Fig.~\ref{fig:neuron_entropy} colors each sample by the sequence’s average entropy. The plot reveals clear clustering of responses by their activated-neuron patterns; importantly, samples within the same cluster do not necessarily share similar entropy values, and samples from different clusters can exhibit similar entropy. This indicates that the activated-neuron patterns encode high-dimensional structure that scalar metrics such as entropy cannot represent.

Taken together, these results suggest that confidence-type scalar metrics are effectively low-dimensional projections of high-dimensional activated-neuron states\footnote{We cannot exhaust all possible activated neuron patterns and their mappings to scalar metrics. In future work, we aim to explore more combinations.}. This observation naturally raises the question: if scalar confidence can guide answer selection, can we exploit richer, higher-dimensional state information to guide selection more effectively? To explore this possibility, in the next section we conduct a deeper analysis of the relationship between neuron activation patterns and response correctness.

\subsection{Preliminary Experiments}
\label{sec:pre_experiment}
To investigate the correlation between neuron activation patterns and model performance, we continue with the same experimental setup, while recording both the correctness of each response and its corresponding set of activated neurons. 

\begin{figure}[tb]
    \centering
    \includegraphics[width=0.98\textwidth]{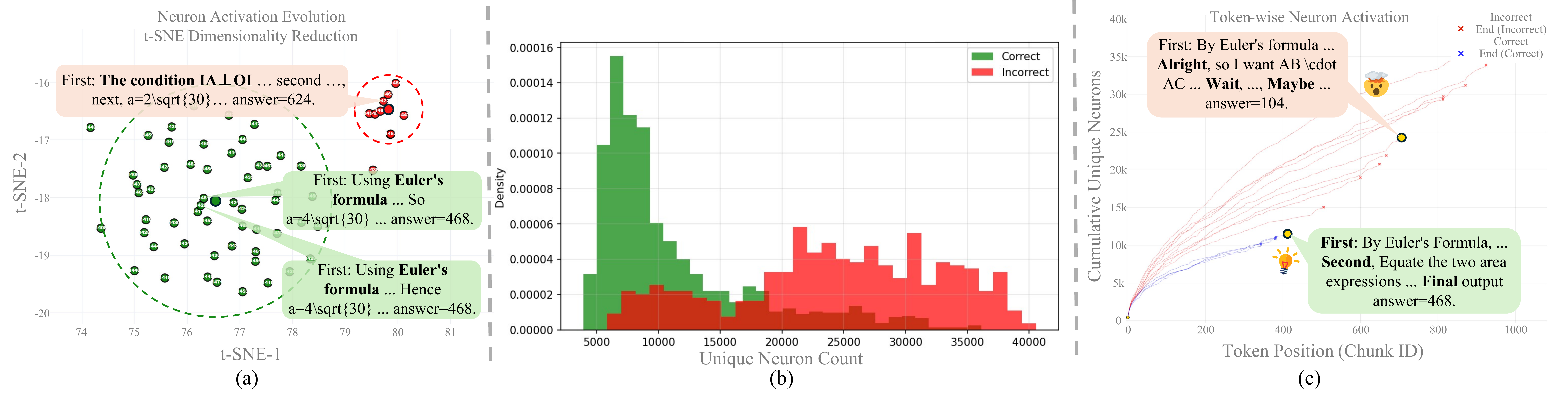}
    \caption{Preliminary AIME24 results. (a) t-SNE of responses to one prompt: center clusters share similar reasoning; outliers diverge. (b) Correct answers activate far fewer neurons than incorrect ones. (c) Token-wise trajectories show incorrect responses repeatedly shift strategies, engaging more neurons. These observations motivate \textbf{Insight 1} and \textbf{Insight 2} presented in Section~\ref{sec:pre_experiment}.}
    \label{fig:preliminaries}
\end{figure}

Figure~\ref{fig:preliminaries}(a) shows the visualization of multiple samples for selected instances, where green points correspond to correct responses and red points to incorrect ones. We observe two clear patterns:  
1) Neuron activation patterns across different responses tend to form natural clusters, indicating a form of consensus;  
2) Incorrect responses tend to lie at the margins of clusters, farther from neighboring samples, whereas correct reasoning trajectories align more tightly with the central distribution. These lead to our first key insight:

\textbf{Insight 1.} By leveraging neuron activation patterns across responses, we can define consensus and identify reasoning trajectories more likely to be correct without requiring text-level matching.

Complementing the clustering result, we compare neuron activations for correct vs. incorrect samples during generation (Fig. \ref{fig:preliminaries}(b,c)). In Fig. (b), correct samples consistently exhibit substantially fewer activated neurons than incorrect ones, consistent with the view that successful trajectories balance exploration and exploitation—reaching answers with fewer trial-and-error steps—whereas failures over-explore. Fig. (c) tracks the growth of unique activated neurons with generated tokens for each sampling; correct trajectories activate fewer neurons throughout. This pattern aligns with the overall statistics and suggests using activation signals to terminate low-quality generations early. These observations lead to our second key insight:

\textbf{Insight 2.} The number of activated neurons in the early stage of generation can serve as a signal to distinguish high-quality answers from low-quality ones. Specifically, high-quality answers tend to activate fewer neurons.

Aligned by generation step (Fig.~\ref{fig:preliminaries}(c)), correct chains activate fewer neurons at matched tokens, indicating a non-length effect and motivating chunk-conditioned early stopping.

Building on these two insights, we can design corresponding algorithms that leverage neuron activation patterns across samples to uncover consensus and identify higher-quality answers.

%% file: chapters_arxiv/method.tex
\section{Methodology}
\label{sec:method}
\begin{figure}[tb]
    \centering
    \includegraphics[width=\textwidth]{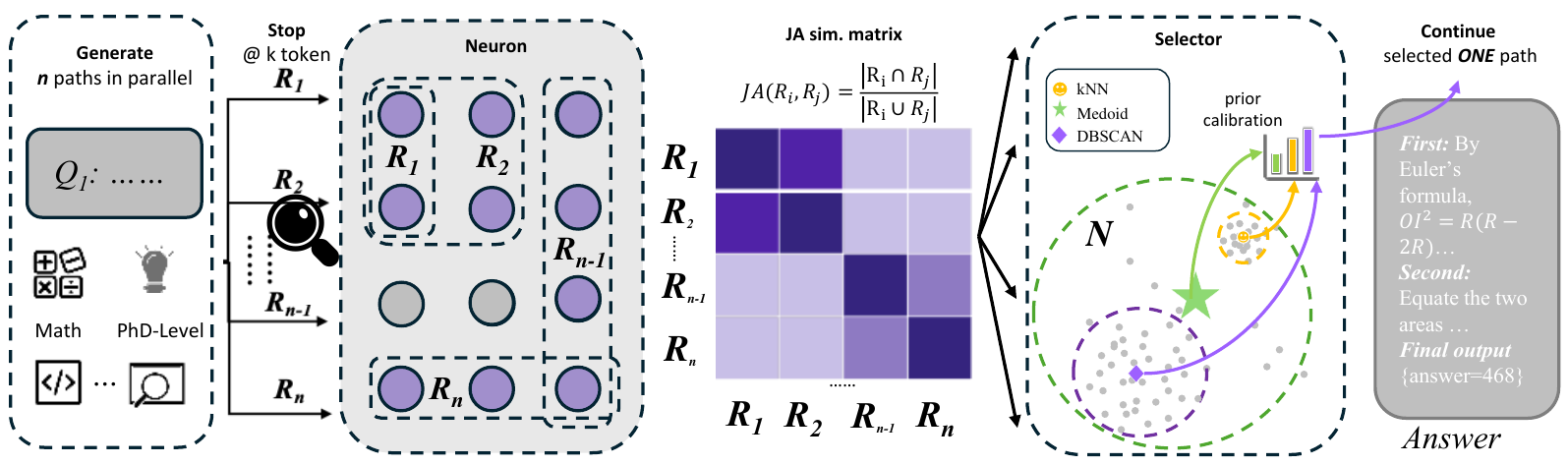}
    \caption{Framework of Neuron Agreement Decoding (NAD). NAD selects high-quality answers by leveraging the consensus of internal neuron activations during the sampling process, without relying on canonical textual outputs. This consistency can be identified using the proposed kNN-based approach, among others. Moreover, this procedure can be applied at an early stage of sequence generation, pruning low-quality responses in advance and reducing token usage.}
    \label{fig:framework}
\end{figure}

Building on the insights from above preliminary experiments~(Section~\ref{sec:pre_experiment}), we aim to operationalize two key observations regarding neuron activations in sampled reasoning trajectories. These observations suggest two complementary strategies for selecting high-quality reasoning trajectories. In the following, we introduce two independent methods that correspond to these insights. Figure~\ref{fig:framework} illustrates the overall framework.

\subsection{Neuron-Agreement Decoding (NAD)}
\label{sec:consensus}

To operationalize the notion of \emph{consensus}, we build directly on \textbf{Insight 1}. Samples of the same input exhibiting similar neuron activation patterns tend to correspond to correct reasoning, while incorrect ones deviate substantially. We capture consensus by exploiting these internal dynamics of the model.

Concretely, for a given input $x$, we generate $n$ sampled trajectories $\{(x,\bm{y}_i)\}_{i=1}^n$, and obtain their activated neuron sets $\{N_\text{activated}(x,\bm{y}_i)\}_{i=1}^n$ as defined in Section~\ref{sec:neuron}. Pairwise similarities among samples, already formalized by the Jaccard index in Section~\ref{sec:pre_experiment}, provide a relational structure over the sampled responses. The resulting consensus matrix $S \in [0,1]^{n \times n}$ captures agreement among samples at the level of neuron activations.

Building upon this representation, our goal is to identify consensus samples --- those most consistent with others in their neuron activations --- and thereby select trajectories that are more likely to be correct. We propose the following structure discovery methods:

\textbf{$k$NN-Agreement.} For each sample $i$, compute the sum of its top-$k$ pairwise similarity scores to other solutions, denoted as $s_i$. Select the solution $\hat{i}$ with the highest $s_i$.

\textbf{Global Medoid.} The medoid denotes the point that minimizes the total distance to all other samples. Under the Jaccard Index metric, this corresponds to maximizing the sum of similarities:
$$
\hat{i}=\arg\max_i\sum\limits_{j=1}^nS_{ij}.
$$

\textbf{DBSCAN.} We apply the clustering algorithm to the distance matrix $D = 1 - S$ to identify clusters. Select the largest cluster $C$, then find the medoid within this cluster:
$$
\hat{i}=\arg\max_i\sum\limits_{p\in C}\sum\limits_{q \in C}S_{pq}.
$$

Building on \textbf{Insight 2}, we propose an alternative selection strategy: since correct reasoning trajectories tend to activate relatively fewer neurons, we choose the trajectory with the fewest activated neurons among the $n$ samples:
$$
\hat{i} = \arg\min_i |N_\text{activated}(x, \pmb{y}_i)|.
$$
This approach does not rely on pairwise similarities between samples; instead, it treats the number of activated neurons in a single sample as a proxy for its quality, allowing us to select trajectories that are more likely to be correct efficiently. We denote this approach as \textbf{MinAct}.

\subsection{Early Stopping Strategy}
\label{sec:early_stop}

Early stopping in parallel sampling improves LLM inference by terminating weak reasoning traces and reallocating compute to stronger ones, reducing redundancy. Based on preliminary experiments, we hypothesize that a trace’s correctness can be predicted early from its neuron activation patterns. We apply these methods to prune low-quality traces. Specifically, for a partial output $\bm{y}_{\le j} = (y_1, y_2, \ldots, y_j)$, we compute the set of neurons activated by $x$ up to the current position $j$ as $N_\text{activated}(x,\bm{y}_{\le j})$ by Eq.(\ref{eq:neuron_chunked}), employ the selection schemes from Section~\ref{sec:consensus}, and resume generation from the selected trace by itself. We set $j$ equal to the chunk size, $B=32$ for the experiments. The choice of the early stopping position will be further discussed in Section~\ref{sec:analysis}.

%% file: chapters_arxiv/experiments.tex
\section{Experiments}
\label{sec:experiment}
\subsection{Setup}
\textbf{Models.} We evaluate NAD on three models: Qwen3-4B-thinking-0527, Qwen3-4B-Instruct-0527~\citep{qwen3technicalreport} and DeepSeek-R1-0528-Qwen3-8B~\citep{deepseekai2025deepseekr1incentivizingreasoningcapability}. For each input, we generate $n=64$ samples with a temperature of $0.6$ and a top-$p$ value of $0.9$.

\textbf{Datasets.} 
We evaluate NAD on two settings: (1) scientific reasoning with canonical answers, including AIME24, AIME25~\citep{aime24a,aime24b,aime25a,aime25b} and GPQA~\citep{rein2024gpqa}); and (2) open‑ended code generation, including LiveCodeBench v5~\citep{jain2024livecodebench}, HumanEval~\citep{chen2021evaluating} and MBPP~\citep{austin2021program}, where majority voting is inapplicable.

\textbf{Protocol \& Baselines.}
Under this protocol we report two baselines: (i) \textbf{Avg@64} (mean accuracy over all $n$ samples); (ii) \textbf{Cons@64} majority vote for tasks with canonical answers (ties count as failure). 
We evaluate under a fixed sampling budget $n=64$ and a low-interaction regime that mirrors deployments where repeated environment calls are expensive; for code, we adopt a single-execution protocol (only the finally selected candidate is executed once).

\begin{table}[htbp]
\centering
\scalebox{0.9}{
\begin{tabular}{cccccccc}
\toprule
\multirow{2}{*}{Model} & \multirow{2}{*}{Method} & \multicolumn{2}{c}{\textbf{Math Reasoning}} & \multicolumn{3}{c}{\textbf{Code Generation}} & \multirow{2}{*}{Avg.}\\
\cmidrule(lr){3-4} \cmidrule(lr){5-7}
& & AIME24+25 & GPQA & HumanEval & LCBv5 & MBPP\\
\midrule
\multirow{6}{*}{Qwen3-4B-Think} 
 & Avg@64 & 74.6 & 66.3 & 96.0 & \textbf{61.9} & 84.6 & 76.7 \\
 & Cons@64 & \textbf{86.7} & \underline{68.2} & -- & -- & -- & --\\
 & \cellcolor{highlight}NAD-kNN & \cellcolor{highlight}\underline{85.0} & \cellcolor{highlight}\textbf{68.7} & \cellcolor{highlight}\textbf{98.2} & \cellcolor{highlight}\underline{61.7} & \cellcolor{highlight}\textbf{86.0} & \cellcolor{highlight}\textbf{79.9}\\
 & \cellcolor{highlight}NAD-Medoid & \cellcolor{highlight}81.7 & \cellcolor{highlight}66.2 & \cellcolor{highlight}97.0 & \cellcolor{highlight}59.3 & \cellcolor{highlight}\underline{85.2} & \cellcolor{highlight}77.9 \\
 & \cellcolor{highlight}NAD-DBSCAN & \cellcolor{highlight}83.4 & \cellcolor{highlight}66.2 & \cellcolor{highlight}\underline{97.6} & \cellcolor{highlight}59.9 & \cellcolor{highlight}85.0 & \cellcolor{highlight}\underline{78.4}\\
 & \cellcolor{highlight}NAD-MinAct & \cellcolor{highlight}\underline{85.0} & \cellcolor{highlight}66.7 & \cellcolor{highlight}92.7 & \cellcolor{highlight}58.1 & \cellcolor{highlight}83.8 & \cellcolor{highlight}77.3\\
 \midrule
 \multirow{6}{*}{R1-Qwen3-8B} 
 & Avg@64 & 70.6 & 58.1 & \textbf{92.1} & \underline{58.5} & 83.7 & 72.6 \\
 & Cons@64 & \textbf{78.3} & \textbf{62.6} & -- & -- & -- & --\\
 & \cellcolor{highlight}NAD-kNN & \cellcolor{highlight}\textbf{78.3} & \cellcolor{highlight}\textbf{62.6} & \cellcolor{highlight}89.6 & \cellcolor{highlight}57.5 & \cellcolor{highlight}\underline{84.4} & \cellcolor{highlight}\textbf{74.5} \\
 & \cellcolor{highlight}NAD-Medoid & \cellcolor{highlight}\underline{75.0} & \cellcolor{highlight}\underline{61.1} & \cellcolor{highlight}\underline{91.5} & \cellcolor{highlight}55.1 & \cellcolor{highlight}\textbf{85.4} & \cellcolor{highlight}\underline{73.6} \\
 & \cellcolor{highlight}NAD-DBSCAN & \cellcolor{highlight}73.3 & \cellcolor{highlight}60.1 & \cellcolor{highlight}90.9 & \cellcolor{highlight}55.7 & \cellcolor{highlight}\textbf{85.4} & \cellcolor{highlight}73.1 \\
 & \cellcolor{highlight}NAD-MinAct & \cellcolor{highlight}\underline{75.0} & \cellcolor{highlight}\underline{61.1} & \cellcolor{highlight}90.2 & \cellcolor{highlight}\textbf{59.3} & \cellcolor{highlight}79.0 & \cellcolor{highlight}72.9\\
 \midrule
 \multirow{6}{*}{Qwen3-4B-Instruct} 
 & Avg@64 & 51.7 & 59.2 & 90.8 & 34.1 & \underline{75.4} & 62.2 \\
 & Cons@64 & \textbf{66.7} & \textbf{61.1} & -- & -- & -- & --\\
 & \cellcolor{highlight}NAD-kNN & \cellcolor{highlight}55.0 & \cellcolor{highlight}\textbf{61.1} & \cellcolor{highlight}\underline{91.5} & \cellcolor{highlight}\textbf{37.1} & \cellcolor{highlight}74.2 & \cellcolor{highlight}63.8 \\
 & \cellcolor{highlight}NAD-Medoid & \cellcolor{highlight}55.0 & \cellcolor{highlight}\textbf{61.1} & \cellcolor{highlight}\textbf{92.1} & \cellcolor{highlight}\underline{36.5} & \cellcolor{highlight}75.2 & \cellcolor{highlight}\underline{64.0} \\
 & \cellcolor{highlight}NAD-DBSCAN & \cellcolor{highlight}60.0 & \cellcolor{highlight}\textbf{61.1} & \cellcolor{highlight}\textbf{92.1} & \cellcolor{highlight}\underline{36.5} & \cellcolor{highlight}\textbf{75.8} & \cellcolor{highlight}\textbf{65.1} \\
 & \cellcolor{highlight}NAD-MinAct & \cellcolor{highlight}\underline{61.7} & \cellcolor{highlight}\underline{60.6} & \cellcolor{highlight}\underline{91.5} & \cellcolor{highlight}31.1 & \cellcolor{highlight}75.0 & \cellcolor{highlight}\underline{64.0} \\
\bottomrule
\end{tabular}
}
\caption{Main results of our experiments. Our methods achieve performance competitive with majority voting and consistently surpass sampling average.}
\label{tab:main_results}
\end{table}

\begin{table}[htbp]
\centering
\scalebox{0.85}{
\begin{tabular}{ccccccc}
\toprule
\multirow{2}{*}{Model} & \multirow{2}{*}{Method} & \multicolumn{2}{c}{\textbf{AIME24+25}} & \multicolumn{2}{c}{\textbf{GPQA}} & \multirow{2}{*}{Avg. Acc.}\\
\cmidrule(lr){3-4} \cmidrule(lr){5-6}
& & Acc. & Token ($\Delta \%$) & Acc. & Token ($\Delta \%$) \\
\midrule
\multirow{5}{*}{Qwen3-4B-Think} 
 & Avg@64 & 74.6 & 55.2 & 66.3 & 102.2  & 70.4\\
 & \cellcolor{highlight}NAD-kNN & \cellcolor{highlight}\underline{80.0} & \cellcolor{highlight}1.3~(-97.6\%) & \cellcolor{highlight}67.2 & \cellcolor{highlight}2.0~(-98.0\%) & \cellcolor{highlight}73.6\\
 & \cellcolor{highlight}NAD-Medoid & \cellcolor{highlight}\textbf{81.7} & \cellcolor{highlight}1.3~(-97.6\%) & \cellcolor{highlight}\textbf{68.2} & \cellcolor{highlight}2.0~(-98.0\%) & \cellcolor{highlight}\textbf{75.0}\\
 & \cellcolor{highlight}NAD-DBSCAN & \cellcolor{highlight}\textbf{81.7} & \cellcolor{highlight}1.3~(-97.6\%) & \cellcolor{highlight}65.7 & \cellcolor{highlight}2.0~(-98.0\%) & \cellcolor{highlight}73.7 \\
 & \cellcolor{highlight}NAD-MinAct & \cellcolor{highlight}\underline{80.0} & \cellcolor{highlight}1.2~(-97.8\%) & \cellcolor{highlight}\underline{67.7} & \cellcolor{highlight}1.8~(-98.2\%) & \cellcolor{highlight}\underline{73.9} \\
 \midrule
 \multirow{5}{*}{R1-Qwen3-8B} 
 & Avg@64 & 70.6 & 48.1 & \textbf{58.1} & 99.6 & 64.4\\
 & \cellcolor{highlight}NAD-kNN & \cellcolor{highlight}\textbf{78.4} & \cellcolor{highlight}1.0~(-97.9\%) & \cellcolor{highlight}\underline{56.1}& \cellcolor{highlight}1.8~(-98.2\%) & \cellcolor{highlight}\underline{67.3} \\
 & \cellcolor{highlight}NAD-Medoid & \cellcolor{highlight}75.0 & \cellcolor{highlight}1.1~(-97.7\%) & \cellcolor{highlight}54.0 & \cellcolor{highlight}1.9~(-98.1\%) & \cellcolor{highlight}64.5\\
 & \cellcolor{highlight}NAD-DBSCAN & \cellcolor{highlight}75.0 & \cellcolor{highlight}1.1~(-97.7\%) & \cellcolor{highlight}54.5 & \cellcolor{highlight}1.9~(-98.1\%) & \cellcolor{highlight}64.8\\
 & \cellcolor{highlight}NAD-MinAct & \cellcolor{highlight}\underline{76.7} & \cellcolor{highlight}0.9~(-98.1\%) & \cellcolor{highlight}\textbf{58.1} & \cellcolor{highlight}1.7~(-98.3\%) & \cellcolor{highlight}\textbf{67.4}\\
 \midrule
 \multirow{5}{*}{Qwen3-4B-Instruct} 
 & Avg@64 & 51.7 & 32.0 & \underline{59.2} & 41.6 & 55.5 \\
 & \cellcolor{highlight}NAD-kNN & \cellcolor{highlight}\underline{55.0} & \cellcolor{highlight}0.4~(-98.8\%) & \cellcolor{highlight}58.6 & \cellcolor{highlight}0.9~(-97.8\%) & \cellcolor{highlight}\underline{56.8}\\
 & \cellcolor{highlight}NAD-Medoid & \cellcolor{highlight}\underline{55.0} & \cellcolor{highlight}0.6~(-98.1\%) & \cellcolor{highlight}56.6 & \cellcolor{highlight}1.0~(-97.6\%) & \cellcolor{highlight}55.8 \\
 & \cellcolor{highlight}NAD-DBSCAN & \cellcolor{highlight}\underline{55.0} & \cellcolor{highlight}0.6~(-98.1\%) & \cellcolor{highlight}56.6 & \cellcolor{highlight}1.0~(-97.6\%) & \cellcolor{highlight}55.8 \\
 & \cellcolor{highlight}NAD-MinAct & \cellcolor{highlight}\textbf{61.7} & \cellcolor{highlight}0.4~(-98.8\%) & \cellcolor{highlight}\textbf{63.1} & \cellcolor{highlight}0.9~(-97.8\%) & \cellcolor{highlight}\textbf{62.4}\\
\bottomrule
\end{tabular}
}
\caption{Accuracy and total token consumption of different methods on scientific reasoning benchmarks after applying early stopping introduced in Section~\ref{sec:early_stop}. Token consumption is reported in millions (M). Our method achieves a two-order-of-magnitude reduction in token usage while maintaining accuracy advantages over random sampling.}
\label{tab:early_stop}
\end{table}

\subsection{Results}
The main results are summarized in Table~\ref{tab:main_results}, which indicate that: 
1) Our approach substantially outperforms baselines in terms of overall performance, with the kNN variant yielding consistently strong performance; 
2) On math reasoning datasets with extractable ground-truth answers, our methods demonstrate clear advantages over sampling average while remaining competitive with majority voting.
On code generation benchmarks, where existing sample-evaluate-ensemble methods are not applicable, our methods still yield performance gains over our curated baseline on most tasks; 
3) Compared to alternative methods that leverage the global structure across $n$ samples, the minimum-activation method relies solely on the number of activated neurons, which limits its effectiveness; nevertheless, it still significantly surpasses baselines.

Table~\ref{tab:early_stop} reports performance changes and token savings on scientific benchmarks under the early-stopping scheme in Section~\ref{sec:early_stop}; code results appear in Appendix Table~\ref{tab:early_stop_code}. Relative to parallel sampling, our method permits stopping after the first chunk, sharply reducing tokens while consistently surpassing random sampling in accuracy. These gains show that early internal neuron activations provide reliable signals of answer quality.

\subsection{Analysis}
\label{sec:analysis}
\begin{figure}[tb]
    \centering
    \includegraphics[width=0.9\textwidth]{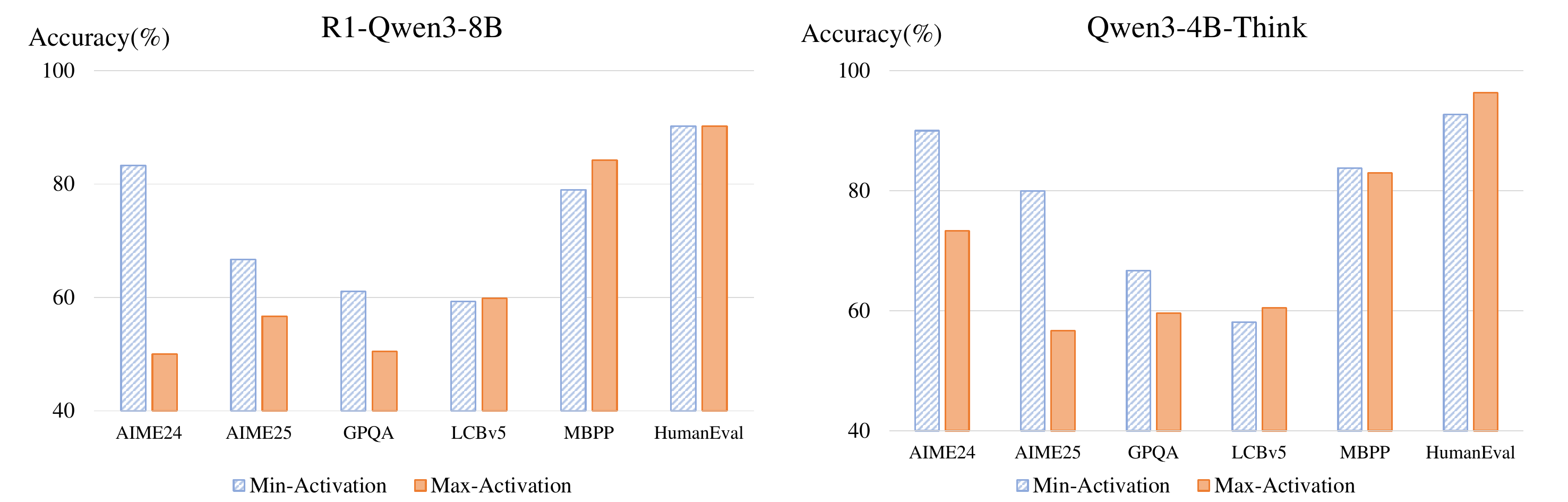}
    \caption{Comparison between minimizing and maximizing activated neurons. On scientific reasoning benchmarks, responses with minimal activated neurons are significantly better; while on coding benchmarks, the performance gap narrows or even reverses.}
    \label{fig:min_max}
\end{figure} 

\begin{figure}[tb]
    \vspace{-7mm}
    \centering
    \includegraphics[width=0.9\textwidth]{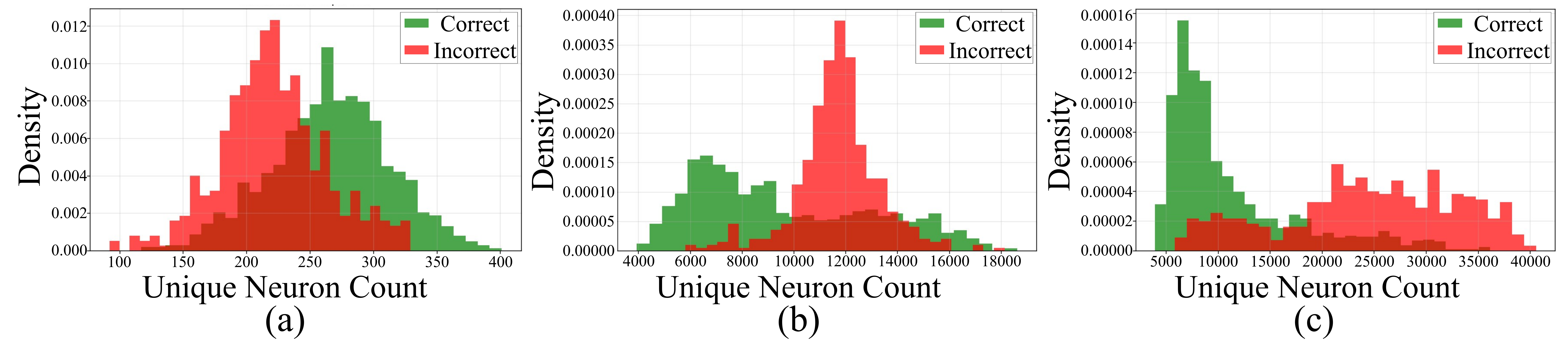}
    \caption{Effect of different top-$k$ settings on the distribution of activated neurons for correct and incorrect responses. From left to right: top-$k$ = 2K, 200K, and no top-$k$. The no-top-$k$ setting achieves the best separation.}
    \label{fig:topk_ablation}
    \vspace{-5mm}
\end{figure}

\textbf{Different implementations for activated neuron computation.} In Section~\ref{sec:neuron}, we aggregate activated neurons by simply taking the union of all chunks, treating them equally. To investigate the effect of this aggregate method, we adopt top-$k$ operation~\citep{cao2025model,wang2025effieval} to merge neurons across sequences. Specifically, we retain only the neurons with the top-$k$ contribution scores, where $k$ ranges from 2K to 200K, and eventually no top-$k$ filtering is applied (our method). The corresponding distributions of activated neurons are shown in Figure~\ref{fig:topk_ablation}.
We observe that as $k$ increases: (1) the distribution of activated neurons for correct samples gradually shifts to the left, while that for incorrect samples shifts to the right; (2) the distribution of incorrect samples becomes increasingly uniform. Overall, the distinction between the two distributions becomes more pronounced.
We hypothesize that when $k$ is small, the activated neurons focus on high-contribution reasoning paths, losing some finer details. As $k$ increases, more detailed information is incorporated, providing a more comprehensive and discriminative view of the model's internal states.

\begin{wrapfigure}{r}{0.45\textwidth}
  \centering
  \includegraphics[width=0.4\textwidth]{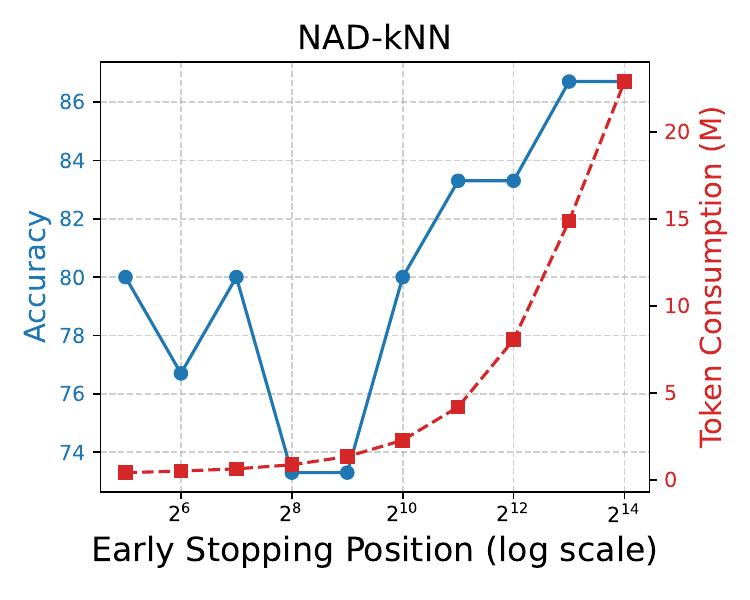} 
  \caption{Accuracy and token consumption as a function of early stopping position. The results show that performance does not monotonically improve as the stopping position is delayed, suggesting that token generation may introduce noise; stopping at 32 tokens achieves relatively good performance.}
  \label{fig:early_stop}
\end{wrapfigure}

\textbf{The Effect of Early Stopping Position.} In the main experiments, we fix the early stopping position at $B=32$. Intuitively, the later the truncation point in generation, the richer the information conveyed by activated neurons. In this section, we investigate how different early stopping positions affect model performance and token consumption. Results across positions ranging from 32 to 16384 are shown in Figure~\ref{fig:early_stop}. Interestingly, we find that later stopping does not necessarily yield better answer quality (For more results, please refer to Figure~\ref{fig:es1}-Figure~\ref{fig:es2} in the Appendix). For example, on the AIME24+25 dataset, the kNN method achieves higher accuracy when stopping at the 4096th token compared to using the full response (86.7 vs. 85.0). This phenomenon may be attributed to noise accumulation and signal dilution: as generation proceeds, additional activations may introduce redundancy or errors that obscure the earlier, more reliable signals, leading to degraded answer selection despite longer reasoning traces.

\textbf{Relationship between neuron activation and performance.} In Section~\ref{sec:pre_experiment}, we observed on \texttt{AIME24} that correct responses activate fewer neurons than incorrect ones. To further examine this phenomenon, we compare it against the opposite strategy: selecting the reasoning trajectory that maximizes neuron activations. The results are reported in Figure~\ref{fig:min_max}. On math and science reasoning benchmarks, our hypothesis is confirmed: responses selected based on minimal activations achieve significantly higher accuracy than those based on maximal activations; whereas on code generation tasks, the difference is much less pronounced, and in some cases, the maximal-activation strategy even outperforms the minimal-activation ones. We suggest that this is because (1) code generation is more open-ended, with multiple valid ways to implement the same functionality, and (2) it inherently requires the model to draw upon a broader range of knowledge, including programming languages, libraries, and domain-specific conventions, which thereby weakens the relationship between neuron activations and performance. We hope that further exploration of this line of research can facilitate extending parallel reasoning to more general domains.

%% file: chapters_arxiv/conclusion.tex
\section{Conclusion}
In this work, we analyze LLM internals via neuron-activation patterns in correct vs. incorrect outputs. Correct outputs activate fewer neurons and align more, a reliable quality signal. We propose Neuron-Agreement Decoding (NAD), selecting responses from internal activations. On math, science, and coding, NAD matches majority voting on well-defined tasks and beats average sampling on open-ended coding. Early pruning cuts tokens up to 99\% without quality loss. These results show that neuron-level signals improve efficiency and reliability, motivating the ensemble decoding based on internal dynamics. 

%% file: chapters_arxiv/appendix.tex
\appendix

\section*{LLM Assistance Disclosure}
\label{sec:llm}
We used large language model (LLM) tools for grammar and wording refinement during manuscript preparation. We also used image-generation tools to create a stylized alpaca illustration to aid reader understanding. All technical content, analyses, and citations were authored, verified, and remain the sole responsibility of the authors.

\section{Limitations}
Despite efficient early selection, open issues remain: (1) \textbf{Selector impact}: we introduced several selectors but did not determine which is most effective under different sampling patterns. (2) \textbf{Storage overhead}: compute cost is small, yet storing neuron activations requires substantial disk and memory; more space-efficient representations of internal dynamics are a key direction. Our comparisons use a fixed budget \(n=64\) and a single-execution protocol for code; approaches with many more samples or multiple executions target a different resource regime. Engineering-wise, storage can be non-trivial, but bitset or bitmap encodings and parallel Jaccard under early stopping (@32 token) keep added overhead small.

\section{Implementation for Threshold Function}
\label{sec:threshold}
In this paper, we adopt a top-$k$ threshold function for key neuron selection, which can be calculated as follows:

\textbf{1. Calculate the highest activations on the $j$-th token $y_j$ in each layer $l$:}
\begin{equation}
\label{eq:tokenwise_activation}
\bm{F}_{jl} = \texttt{topk}(\bm{A}(y_j,l), 64),
\end{equation}
where $\bm{A}(y_j, l)\in \mathbb{R}^{N}$ denotes the contribution score matrix on token $y_j$ in layer $l$, with 
$[\bm{A}(y_j,l)]_{i} = f_\text{neuron}(i, l, y_j \mid x \oplus y_{<j})$. 
Here, $\texttt{topk}(\bm{A}, k)$ returns the $k$ largest values in $\bm{A}$.

\textbf{2. Find the threshold by aggregating activations across the sequence:}
\begin{equation}
\eta(j, k) = \texttt{min}\{\texttt{topk}([\bm{F}_{j1}; \bm{F}_{j2}; \dots; \bm{F}_{jl}], k)\}.
\end{equation}

In all experiments, we set $k=500$. Note that the threshold is equivalent to taking the top-$k$ across all activation values on token $y_j$ when $64$ in Eq.~(\ref{eq:tokenwise_activation}) 
is scaled up to $N$. We choose to use $64$ instead of $N$ for computational efficiency considerations.
This token‑level thresholding is always applied in our main method. Unless otherwise noted (Sec. 5.3), we do not apply any sequence‑level global top‑k across tokens; when we do, it is clearly marked as an ablation. 

\section{Detailed Experiment Results}
\begin{table}[htbp]
\centering
\scalebox{0.75}{
\begin{tabular}{ccccccccc}
\toprule
\multirow{2}{*}{Model} & \multirow{2}{*}{Method} & \multicolumn{2}{c}{HumanEval} & \multicolumn{2}{c}{LCBv5} & \multicolumn{2}{c}{MBPP} & \multirow{2}{*}{Avg. Acc.}\\
\cmidrule(lr){3-4} \cmidrule(lr){5-6} \cmidrule(lr){7-8}
& & Acc. & Token ($\Delta \%$) & Acc. & Token ($\Delta \%$) & Acc. & Token ($\Delta \%$) \\
\midrule
\multirow{5}{*}{Qwen3-4B-Think}
 & Avg@64 & \underline{97.0} & 52.2 & 58.7 & 191.9 & \textbf{85.6} & 169.8 & 80.4\\
 &\cellcolor{highlight}NAD-kNN & \cellcolor{highlight}\textbf{97.6} & \cellcolor{highlight}1.1(-97.9\%) & \cellcolor{highlight}\underline{63.5} & \cellcolor{highlight}3.2(-98.3\%) & \cellcolor{highlight}\underline{85.2} & \cellcolor{highlight}3.5(-98.0\%) & \cellcolor{highlight}\underline{82.1}\\
 & \cellcolor{highlight}NAD-Medoid & \cellcolor{highlight}\textbf{97.6} & \cellcolor{highlight}1.1(-97.9\%) & \cellcolor{highlight}59.9 & \cellcolor{highlight}3.3(-98.3\%) & \cellcolor{highlight}85.0 & \cellcolor{highlight}3.5(-98.0\%) & \cellcolor{highlight}80.8\\
 & \cellcolor{highlight}NAD-DBSCAN & \cellcolor{highlight}\textbf{97.6} & \cellcolor{highlight}1.1(-97.9\%) & \cellcolor{highlight}61.1 & \cellcolor{highlight}3.3(-98.3\%) & \cellcolor{highlight}\underline{85.2} & \cellcolor{highlight}3.6 (-97.9\%) & \cellcolor{highlight}81.3\\
 &\cellcolor{highlight}NAD-MinAct & \cellcolor{highlight}96.3 & \cellcolor{highlight}1.0(-98.1\%) & \cellcolor{highlight}\textbf{65.9} & \cellcolor{highlight}3.2(-98.3\%) & \cellcolor{highlight}\underline{85.2} & \cellcolor{highlight}3.1(-98.2\%) & \cellcolor{highlight}\textbf{82.4}\\
 \midrule

 \multirow{5}{*}{R1-Qwen3-8B}
 & Avg@64 & 92.1 & 49.3 & \underline{61.1} &  190.4& \textbf{84.6} & 171.6 & \textbf{79.3}\\
 & \cellcolor{highlight}NAD-kNN    & \cellcolor{highlight}\textbf{94.5} & \cellcolor{highlight}0.9(-98.2\%) & \cellcolor{highlight}56.9 & \cellcolor{highlight}3.2(-98.3\%) & \cellcolor{highlight}83.2 & \cellcolor{highlight}3.3(-98.1\%) & \cellcolor{highlight}78.2 \\
 & \cellcolor{highlight}NAD-Medoid & \cellcolor{highlight}91.5 & \cellcolor{highlight}1.0(-98.0\%) & \cellcolor{highlight}58.7 & \cellcolor{highlight}3.2(-98.3\%) & \cellcolor{highlight}\underline{84.0} & \cellcolor{highlight}3.5(-98.0\%) & \cellcolor{highlight}78.1\\
 & \cellcolor{highlight}NAD-DBSCAN & \cellcolor{highlight}91.5 & \cellcolor{highlight}1.0(-98.0\%) & \cellcolor{highlight}\textbf{61.7} & \cellcolor{highlight}3.2(-98.3\%) & \cellcolor{highlight}83.0 & \cellcolor{highlight}3.5(-98.0\%) & \cellcolor{highlight}\underline{78.7}\\
 & \cellcolor{highlight}NAD-MinAct & \cellcolor{highlight}\underline{93.9} & \cellcolor{highlight}0.9(-98.2\%) & \cellcolor{highlight}58.7 & \cellcolor{highlight}2.9(-98.4\%) & \cellcolor{highlight}82.4 & \cellcolor{highlight}3.0(-98.2\%) & \cellcolor{highlight}78.3 \\
 \midrule
 \multirow{5}{*}{Qwen3-4B-Instruct}
 & Avg@64 & 89.6 & 6.0 & 31.1 & 26.6 &  74.9 & 37.1 & 65.2\\
 & \cellcolor{highlight}NAD-kNN    & \cellcolor{highlight}90.2 & \cellcolor{highlight}0.4(-93.3\%) & \cellcolor{highlight}\underline{35.9} & \cellcolor{highlight}0.6(-97.7\%) & \cellcolor{highlight}73.9 & \cellcolor{highlight}1.4(-96.2\%) & \cellcolor{highlight}66.7\\
 & \cellcolor{highlight}NAD-Medoid & \cellcolor{highlight}\underline{90.9} & \cellcolor{highlight}0.4(-93.3\%) & \cellcolor{highlight}\textbf{36.5} & \cellcolor{highlight}0.7(-97.4\%) & \cellcolor{highlight}74.7 & \cellcolor{highlight}1.5(-96.0\%) & \cellcolor{highlight}\underline{67.4}\\
 & \cellcolor{highlight}NAD-DBSCAN & \cellcolor{highlight}\underline{90.9} & \cellcolor{highlight}0.4(-93.3\%) & \cellcolor{highlight}\underline{35.9} & \cellcolor{highlight}0.7(-97.4\%) & \cellcolor{highlight}\underline{76.2} & \cellcolor{highlight}1.5(-96.0\%) & \cellcolor{highlight}\textbf{67.7}\\
 & \cellcolor{highlight}NAD-MinAct & \cellcolor{highlight}\textbf{92.1} & \cellcolor{highlight}0.4(-93.3\%) & \cellcolor{highlight}31.7 & \cellcolor{highlight}0.6(-97.7\%) & \cellcolor{highlight}\textbf{77.0} & \cellcolor{highlight}1.4(-96.2\%) & \cellcolor{highlight}66.9\\
\bottomrule

\end{tabular}
}
\caption{Accuracy and total token consumption of different methods on code benchmarks after applying early stopping introduced in Section~\ref{sec:early_stop}. Token consumption is reported in millions (M). Our method achieves a two-order-of-magnitude reduction in token usage while maintaining accuracy advantages over random sampling.}
\label{tab:early_stop_code}
\end{table}

\begin{figure}[tb]
    \centering
    \includegraphics[width=0.95\textwidth]{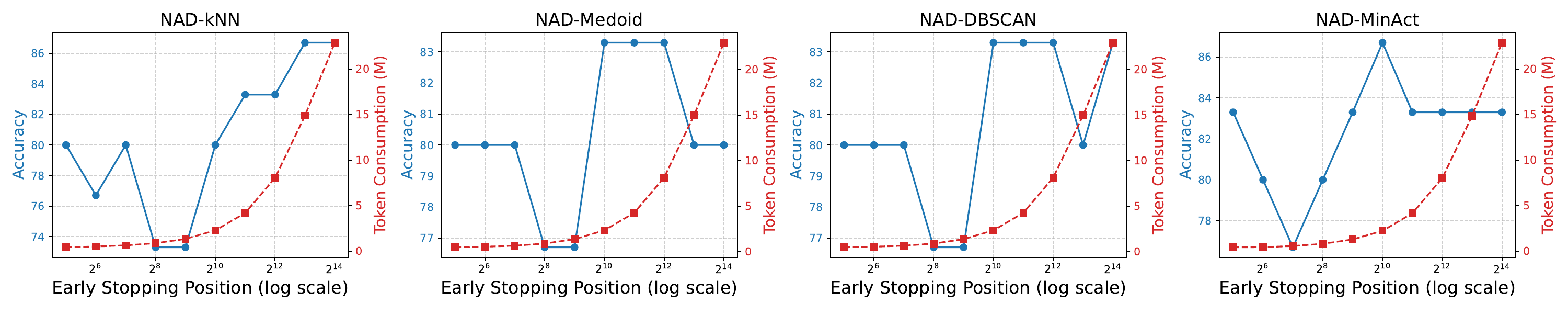}
    \caption{Accuracy and token consumption as a function of early stopping position of \texttt{R1-Qwen3-8B} on \texttt{AIME24}.}
    \label{fig:es1}
\end{figure} 

\begin{figure}[tb]
    \centering
    \includegraphics[width=0.95\textwidth]{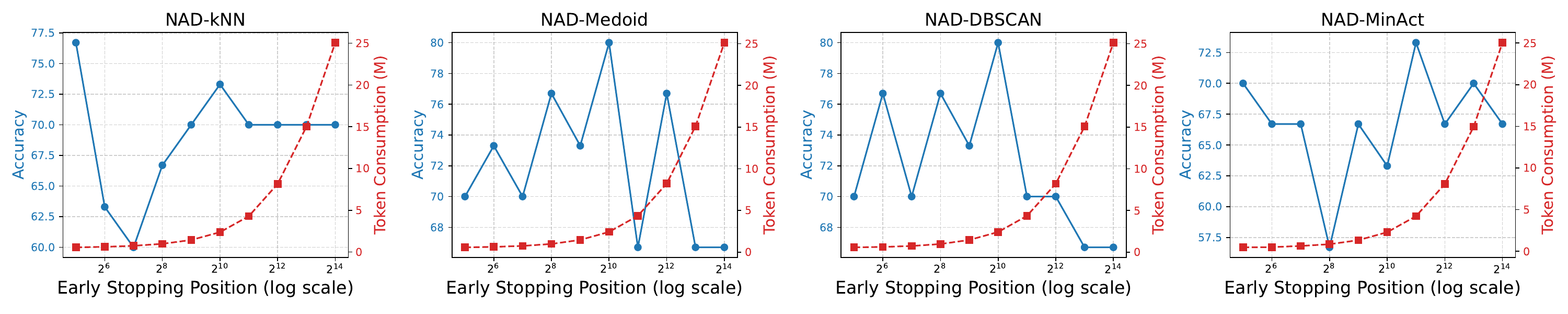}
    \caption{Accuracy and token consumption as a function of early stopping position of \texttt{R1-Qwen3-8B} on \texttt{AIME25}.}
    \label{fig:es2}
\end{figure}